\documentclass{article} 
\usepackage{iclr2027_conference,times}
\iclrfinalcopy

\usepackage{amsmath,amsfonts,bm}

\def\eqref#1{equation~\ref{#1}}

\def\1{\bm{1}}

\DeclareMathAlphabet{\mathsfit}{\encodingdefault}{\sfdefault}{m}{sl}
\SetMathAlphabet{\mathsfit}{bold}{\encodingdefault}{\sfdefault}{bx}{n}

\usepackage{hyperref}
\usepackage{url}
\usepackage{wrapfig}
\usepackage{subcaption}
\usepackage{graphicx}
\usepackage{cleveref}
\usepackage{booktabs}
\usepackage{multirow}
\usepackage{xcolor}
\usepackage{colortbl} 
\usepackage{pifont}   
\usepackage{amssymb}

\usepackage{bm}
\usepackage{algorithm}
\usepackage{algorithmic}
\usepackage{stfloats}

\usepackage{booktabs}
\usepackage{graphicx}       
\usepackage{multirow}       
\usepackage{threeparttable}
\usepackage{pifont}

\usepackage{amsmath}

\usepackage{booktabs}
\usepackage{threeparttable}
\usepackage{siunitx}

\Crefname{figure}{Fig.}{Figs.}

\Crefname{table}{Tab.}{Tabs.}

\Crefname{section}{Sec.}{Secs.}
\Crefname{equation}{Eq.}{Eqs.}

\title{UniRect-CoT: Enhancing Generation in Unified Multimodal Models via Reflective Rectification with Inherent Understanding}

\author{
\normalfont
\textbf{Yibo Jiang}$^{1}$\thanks{Equal contribution.} \quad
\textbf{Tao Wu}$^{2,*}$ \quad
\textbf{Rui Jiang}$^{2}$ \quad
\textbf{Yehao Lu}$^{2}$ \\[0.3em]
\textbf{Chaoxiang Cai}$^{1}$ \quad
\textbf{Zequn Qin}$^{1}$\thanks{
Corresponding authors: Zequn Qin and Xi Li.
Contact: \texttt{xilizju@zju.edu.cn} (Xi Li).
} \quad
\textbf{Xi Li}$^{2,\dagger}$ \\[0.6em]
\normalfont
$^{1}$\,School of Software Technology, Zhejiang University \\
\normalfont
$^{2}$\,College of Computer Science and Technology, Zhejiang University
}

\hypersetup{hidelinks}

\begin{document}

\maketitle

\fancyhead{}
\fancyhead[L]{Preprint}

\begin{abstract}
Unified Multimodal Models (UMMs) aim to integrate visual understanding and generation within a single structure.
However, these models exhibit a notable capability mismatch, where their understanding capability often outperforms their generation capability.
This mismatch suggests that the model's rich internal knowledge, while effective for understanding tasks, is not fully utilized during generation.
To address this, we draw inspiration from the human ``Thinking-While-Drawing'' paradigm, where humans continuously reflect on intermediate results and revise them according to their understanding of the intended target.
In this paper, we propose UniRect-CoT, a training-free unified rectification chain-of-thought framework.
We regard the multi-step denoising process in UMMs as an intrinsic visual reasoning process, whose intermediate states provide opportunities for continuous reflection and rectification.
By leveraging the UMM's inherent understanding to reflect on intermediate results in light of the target instruction and rectify them accordingly, our framework forms a reflective chain of thought within a single denoising trajectory, enabling the model's internal knowledge to guide generation throughout the process.
Specifically, the UMM generates a state-conditioned reflection, whose autoregressive negative log-likelihood under the same model defines a model-native differentiable semantic loss that translates textual reflection into gradients for latent rectification.
Extensive experiments demonstrate that UniRect-CoT generalizes across existing flow-based UMMs, consistently enhancing overall text-to-image generation performance.
\end{abstract}
\section{Introduction}
\label{sec:intro}

Building upon the success of Large Language Models (LLMs)~\cite{guo2025deepseek,team2025kimi}, Multimodal Large Language Models (MLLMs) have been developed to incorporate visual representations into language-centric frameworks.
Recently, Unified Multimodal Models (UMMs) have emerged to integrate ``understanding'' and ``generation'' within a single structure.
Such a unified structure creates the potential to leverage the model's understanding capability during generation.

\begin{figure*}[t]
    \centering
    \includegraphics[width=\linewidth]{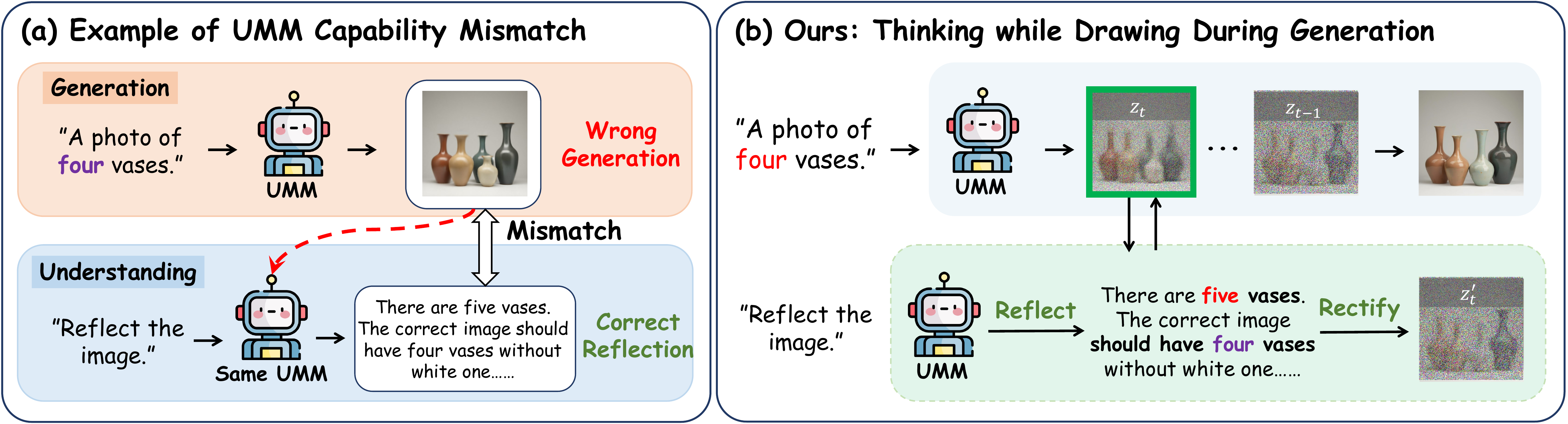}
    \caption{
    \textbf{Capability mismatch in UMMs and overview of UniRect-CoT.}
    (a) A UMM may fail to follow the generation instruction while its understanding branch can identify the error and describe the desired correction.
    (b) UniRect-CoT realizes ``Thinking-While-Drawing'' by interleaving state-conditioned reflection and rectification within the same denoising trajectory.
    }
    \label{fig:intro}
    \vspace{-3mm}
\end{figure*}

Although UMMs achieve structural unification of understanding and generation through shared parameters, they exhibit a notable capability mismatch, where the understanding capability of UMMs often outperforms their generation capability~\cite{xie2025reconstruction}.
As shown in \Cref{fig:intro}(a), given the instruction ``A photo of four vases'', the UMM incorrectly generates five vases, while its understanding branch correctly identifies the extra object and describes the desired correction. 
This example suggests that, while UMMs effectively leverage their internal knowledge for understanding, this same knowledge is not fully used during generation.
Therefore, a natural question arises: Can we use the UMM's inherent understanding capability to guide generation?

To achieve this, we draw inspiration from the human creation process of ``Thinking-While-Drawing.''
Specifically, given an instruction, humans continuously reflect on intermediate results by comparing the current result with the intended target and revise them accordingly.
This reflective process allows existing knowledge to be used throughout creation.
In contrast, current UMMs lack an explicit reflection mechanism for generation.
Fortunately, the multi-step denoising process in UMMs can be regarded as an intrinsic visual reasoning process, naturally producing intermediate results that resemble the human creation process.
This offers a valuable opportunity to leverage the UMM's understanding capability to iteratively reflect on and rectify intermediate generation results during denoising, forming a chain of thought that guides generation.

Based on these insights, we propose UniRect-CoT (\Cref{fig:intro}(b)), a training-free unified rectification chain-of-thought framework that interleaves reflection and rectification within the same denoising trajectory.
At each step of this chain, the UMM first inspects the current intermediate result, reflects on its discrepancies with the user instruction, and then uses this reflection to rectify the intermediate state.
Specifically, at selected timesteps, we map the noisy intermediate state to an estimated clean image that can be inspected by the UMM's understanding branch.
Based on the estimated image and the user instruction, the UMM produces a state-conditioned reflective description of what the generated image should depict to satisfy the instruction.
Using the estimated image as visual context, we teacher-force the reflective description through the same UMM and use its autoregressive negative log-likelihood as a model-native differentiable semantic loss.
Backpropagating this loss to the intermediate state provides gradients for rectification without external semantic supervision.
To improve the reliability of rectification, we further introduce Greedy Iterative Trajectory Optimization (GITO), which explores successive rectified candidates and selects among them according to the user instruction.
By repeatedly applying this process, UniRect-CoT enables the UMM to leverage its understanding capability for continuous reflection and rectification during generation, realizing ``Thinking-While-Drawing.''


The key contributions of this work are summarized as follows:
\begin{itemize}

\item We propose UniRect-CoT, a training-free unified rectification chain-
of-thought framework that realizes ``Thinking-While-Drawing'' by interleaving state-conditioned reflection and rectification within a single denoising trajectory.

\item We develop a model-native differentiable semantic loss from the UMM's autoregressive likelihood, converting discrete textual reflection into gradients for latent rectification without additional training or external semantic supervision.

\item Extensive experiments across four flow-based UMMs and two text-to-image benchmarks demonstrate the effectiveness and generality of UniRect-CoT.

\end{itemize}
\section{Related Work}
\label{sec:related_work}

\subsection{Unified Multimodal Models}

Unified Multimodal Models (UMMs) integrate multimodal understanding and generation within a single framework~\cite{zhang2025unifiedmultimodalunderstandinggeneration}.
Existing UMMs follow two paradigms: autoregressive (AR) and AR+Diffusion.
\textit{The AR paradigm} includes next-token prediction architectures~\cite{team2024chameleon,chen2025janus,lwm,qu2024tokenflow}, which unify the prediction of textual and visual tokens, as well as Masked Autoregressive (MAR) models~\cite{li2024autoregressive} and their variants~\cite{wu2025harmon}, which support direct generation without vector quantization.
\textit{The AR+Diffusion paradigm} combines Transformers with diffusion or flow-matching objectives through architectural designs.
Serial configurations~\cite{lin2025uniworld,wu2025qwen,wu2025omnigen2explorationadvancedmultimodal} use AR representations to condition the diffusion generation process;
parallel mechanisms~\cite{shi2024lmfusion,bagel,wang2025lightfusionlightweighteddoublefusion} connect understanding and generation through shared intermediate features;
and unified structures~\cite{zhou2025transfusion,xie2025show} directly incorporate diffusion or flow-matching heads into a shared Transformer backbone.
In this paper, we focus on AR+Diffusion UMMs, as our method leverages the UMM's understanding capability to rectify its denoising trajectory.

\subsection{Test-Time Scaling for Multimodal Generation}

Test-time scaling improves multimodal generation by allocating additional computation at inference time.
Recent methods introduce chain-of-thought (CoT) reasoning through textual or interleaved multimodal representations~\cite{gu2025improving,qin2025uni}, while others perform multi-round generation and refinement by revising textual conditions based on previously generated outputs~\cite{khan2025testtimepromptrefinementtexttoimage,lyu2025understandingingenerationreinforcinggenerativecapability}.
These methods mainly reason before generation or refine across multiple generation rounds.
In contrast, UniRect-CoT performs state-conditioned reflection and rectification within the same denoising trajectory, allowing the reflection to adapt to intermediate generation results.
Gradient-based guidance steers diffusion or flow-based sampling through differentiable objectives~\cite{dhariwal2021diffusion,liu2023more,song2023loss,bansal2023universal,yu2023freedom}.
These methods establish a general mechanism for modifying the generation trajectory at inference time.
UniRect-CoT builds on this mechanism but differs in how the semantic loss is constructed: we derive a model-native differentiable semantic loss from the UMM's own reflection and autoregressive likelihood, enabling textual reflection to guide trajectory rectification without an external semantic oracle.
\section{Preliminaries}
\label{sec:preliminary}


\noindent\textbf{Flow Matching}.
Flow Matching~\cite{lipman2022flow} has emerged as the successor to DDPMs in text-to-image synthesis, distinguished by its superior sampling efficiency.
Standard implementations typically utilize a pre-trained autoencoder (comprising encoder $\mathcal{E}$ and decoder $\mathcal{D}$) to map input images $x$ into a latent representation $z_0 = \mathcal{E}(x)$.
Let $z_0$ denote a clean latent, $\epsilon\sim\mathcal N(0,I)$ Gaussian noise, and $c$ a text prompt.
The generative process is modeled as a deterministic Ordinary Differential Equation (ODE), $\frac{dz_t}{dt} = v(z_t, t)$ for $t \in [0,1]$, which defines a continuous trajectory from Gaussian noise $\epsilon$ to data $z_0$.
This trajectory is typically parameterized via linear interpolation: $z_t = (1 - t)z_0 + t \epsilon$.
To enable generation, a velocity network $v_\theta$ is trained to regress the vector field by minimizing the matching objective:
\begin{equation} \label{eq:diff_loss}
\mathcal{L}_{diff} = \mathbb{E}_{t, z_0, \epsilon} \| (\epsilon - z_0) - v_\theta(z_t, t, c)\|^2.
\end{equation}

To bridge intermediate noisy states and the clean data, {Look-Ahead Estimation} leverages a core property of linear flow parameterization that $z_0$ can be estimated from any $z_t$.
Similar to Tweedie's formula in the context of diffusion models, the estimated clean latent $\hat{z}_{0|t}$ can be derived directly from the current state and the predicted velocity:
\begin{equation} \label{eq:z0_pred}
\hat{z}_{0|t} = z_t - t \cdot v_\theta(z_t, t, c).
\end{equation}
This estimation serves as a bridge for applying operations defined on clean data to noisy states.

\noindent\textbf{Gradient-Based Guidance for Flow Models.}
Gradient-based guidance steers diffusion or flow-based generation using gradients from differentiable objectives~\cite{dhariwal2021diffusion,bansal2023universal,chung2022diffusion,song2023loss,yu2023freedom}.
Given a target condition $y$ and a differentiable objective $\mathcal{L}(z_0,y)$, the conditional distribution can be expressed as
$p_0(z_0|y)\propto p_0(z_0)\exp(-\mathcal{L}(z_0,y))$.
For flow-based generation, the corresponding conditional gradient modifies the velocity as
\[
\hat v(z_t,t,c)
=
v_\theta(z_t,t,c)
-
\gamma\nabla_{z_t}\log p_t(y|z_t),
\]
where $\gamma$ controls the guidance strength.
Since the exact conditional gradient is generally intractable, training-free guidance commonly approximates it by evaluating the differentiable objective on a predicted clean sample.
Using the look-ahead estimate in Eq.~\ref{eq:z0_pred}, the guidance gradient can be approximated as
\begin{equation}
\label{eq:flow_guidance}
\nabla_{z_t}\log p_t(y|z_t)
\approx
-\nabla_{z_t}
\mathcal{L}(\hat z_{0|t},y).
\end{equation}
This formulation shows that a differentiable loss defined on the predicted clean sample can provide gradients to rectify the sampling trajectory.
In UniRect-CoT, we use this mechanism as the optimization step for applying reflection to the denoising trajectory.
Our key design in this step lies in the construction of the loss: we derive a model-native differentiable semantic loss from the UMM's state-conditioned reflection and autoregressive likelihood.
\section{Methods}
\label{sec:methods}

\begin{figure*}[t]
    \centering
    \setlength{\abovecaptionskip}{1.5pt}
    \includegraphics[width=\linewidth]{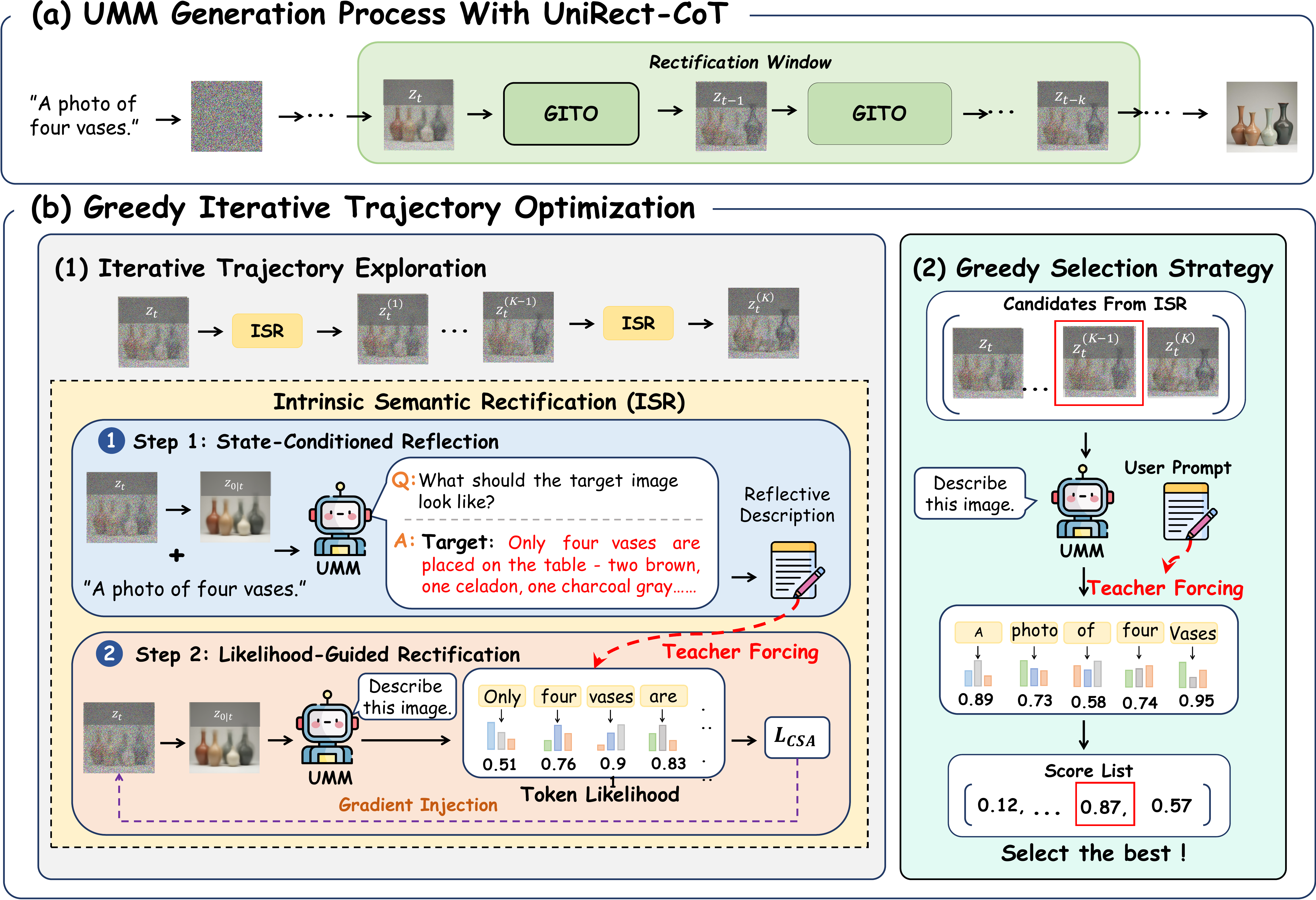}
    \caption{
    \textbf{Overall pipeline of UniRect-CoT.}
    (a) Reflection and rectification are interleaved with denoising within a selected rectification window.
    (b) At each rectification timestep, GITO performs iterative trajectory exploration and greedy selection.
    ISR combines state-conditioned reflection with likelihood-guided rectification. The reflective target is generated in the first ISR update and reused by subsequent updates at the same timestep.
    The resulting candidates, including the original state, are evaluated using the user instruction to determine the state retained for subsequent denoising.
    }
    \label{fig:pipeline}
    \vspace{-3mm}
\end{figure*}

In this section, we present UniRect-CoT(\Cref{fig:pipeline}), a training-free framework that realizes ``Thinking-While-Drawing'' by leveraging the UMM's inherent understanding to reflect on and rectify intermediate results within a single denoising trajectory.
We first evaluate the UMMs' ability to judge whether intermediate generation results satisfy the user instruction, and then describe how successive reflections form a chain along the denoising trajectory in \Cref{subsec:reflective_states}.
Then, in \Cref{subsec:self_reflective}, we introduce Intrinsic Semantic Rectification (ISR), which first generates state-conditioned reflection and then converts its corrective target into a model-native differentiable semantic loss for latent rectification.
Finally, \Cref{subsec:selection_strategy} details our Greedy Iterative Trajectory Optimization (GITO), which explores candidate states through iterative ISR updates and selects the state for subsequent denoising according to the original user instruction.

\subsection{Reflective Reasoning Along the Denoising Trajectory}
\label{subsec:reflective_states}

To enable reflection during generation, we first make intermediate denoising states accessible to the UMM's understanding branch.
For a latent state $z$ at flow time $t$, Look-Ahead Estimation gives
\begin{equation}
\label{eq:reflective_state}
    \hat{z}_{0|t}(z)
    =
    z-t\,v_\theta(z,t,c_{\mathrm{user}}),
    \qquad
    \hat{x}_{0|t}(z)
    =
    \mathcal{D}\!\left(\hat{z}_{0|t}(z)\right).
\end{equation}

\setlength{\intextsep}{2pt}
\begin{wrapfigure}{l}{0.45\columnwidth}
    \centering
    \setlength{\belowcaptionskip}{0pt}
    \setlength{\abovecaptionskip}{1.5pt}
    \includegraphics[width=\linewidth]{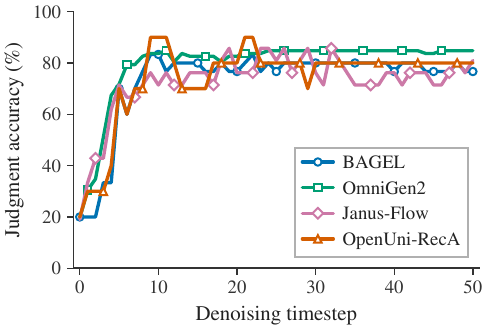}
    \caption{
    \textbf{Judgment accuracy across denoising timesteps.}
    We measure how often each UMM's judgment of whether a look-ahead image satisfies the user instruction agrees with the GenEval correctness label for the same image.
    }
    \label{fig:reflection_accuracy}
\end{wrapfigure}
GenEval~\cite{ghosh2023genevalobjectfocusedframeworkevaluating} is a widely used text-to-image benchmark with explicit correctness criteria for each test case.
To assess whether intermediate denoising states contain sufficient semantic information for evaluation, we apply look-ahead estimation at different timesteps along each trajectory and ask the UMM's understanding branch whether the estimated images satisfy the user instruction.
We measure judgment accuracy at each timestep by comparing the UMM's judgments with the GenEval correctness labels computed on the same images.
As shown in \Cref{fig:reflection_accuracy}, all four UMMs achieve high judgment accuracy well before generation is complete.
This observation motivates using the UMM's understanding branch to assess intermediate results with respect to the original user instruction during denoising.

We propose to interleave reflection and rectification with denoising to iteratively refine intermediate results.
At selected denoising steps, the UMM reflects on and rectifies the current intermediate result, allowing its inherent understanding to guide generation.
As denoising progresses, subsequent reflections are conditioned on intermediate results shaped by preceding rectifications.
This alternation links successive reflections through the evolving visual state, naturally forming a reflective chain of thought within a single denoising trajectory and realizing ``Thinking-While-Drawing.''

\subsection{Intrinsic Semantic Rectification}
\label{subsec:self_reflective}

To operationalize this reflective process, we introduce Intrinsic Semantic Rectification (ISR), which aligns the intermediate result with a reflective target derived from the UMM's own understanding.
The ISR consists of two steps: state-conditioned reflection and likelihood-guided rectification.

\noindent\textbf{Step 1: State-Conditioned Reflection.}
The user instruction specifies the desired content, while state-conditioned reflection identifies discrepancies in the current intermediate result that require rectification.
For example, when the instruction specifies ``four vases'' but the estimated image contains five, the UMM can identify the extra vase and describe the desired four-vase image.
To obtain such feedback, we provide the estimated image $\hat{x}_{0|t}(z_t)$, a reflection prompt $c_{\mathrm{ref}}$, and the user instruction $c_{\mathrm{user}}$ to the UMM's understanding branch:
\begin{equation}
\label{eq:state_reflection}
    r_t
    =
    \mathcal{M}_{\mathrm{und}}
    \!\left(
        \hat{x}_{0|t}(z_t),
        [c_{\mathrm{ref}};c_{\mathrm{user}}]
    \right).
\end{equation}
From the resulting reflection $r_t$, we extract the reflective target $c_{\mathrm{target}}^{t}$ that describes the desired visual content after rectification.

\noindent\textbf{Step 2: Likelihood-Guided Rectification.}
The reflective target describes the desired visual content, but its discrete textual form cannot directly rectify the latent state.
To construct a differentiable objective, we teacher-force $c_{\mathrm{target}}^{t}=(y_1,\ldots,y_L)$ through the same UMM, using the estimated image as visual context.
We define the \textit{Cyclic Semantic Alignment} (CSA) loss as the average autoregressive negative log-likelihood of the reflective target:
\begin{equation}
\label{eq:ce_loss}
    \mathcal{L}_{\mathrm{CSA}}
    \left(z;c_{\mathrm{target}}^{t}\right)
    =
    -\frac{1}{L}
    \sum_{i=1}^{L}
    \log P_\theta
    \!\left(
        y_i
        \mid
        y_{<i},
        \hat{x}_{0|t}(z)
    \right).
\end{equation}
This model-native loss encourages the estimated image to align with the reflective target under the UMM's own understanding, without external supervision.
To minimize the alignment objective, we employ a gradient injection mechanism.
At each selected denoising step, we treat the latent state $z_t$ as the optimization variable while keeping all model parameters and the reflective target fixed.
Differentiating the CSA loss with respect to $z_t$ gives the rectification gradient:
\begin{equation}
\label{eq:latent_gradient}
\begin{aligned}
    g_t
    &=
    \nabla_{z_t}
    \mathcal{L}_{\mathrm{CSA}}
    \left(z_t;c_{\mathrm{target}}^{t}\right)
\end{aligned}
\end{equation}
The gradient is computed through the UMM's understanding branch and propagated through the decoder and generation branch to the intermediate latent state.
To limit excessive perturbations to the denoising trajectory, we apply $\ell_2$-norm gradient clipping:
\begin{equation}
\label{eq:grad_clip}
    \hat{g}_t
    =
    \begin{cases}
        \displaystyle
        \delta\frac{g_t}{\|g_t\|_2},
        & \text{if } \|g_t\|_2>\delta, \\[4pt]
        g_t,
        & \text{otherwise},
    \end{cases}
\end{equation}
where $\delta$ denotes the clipping threshold.
Finally, we update the latent state using the clipped gradient:
\begin{equation}
\label{eq:inject_grad}
    z_t
    \leftarrow
    z_t-\eta\,\hat{g}_t,
\end{equation}
where $\eta$ controls the rectification strength.

\subsection{Greedy Iterative Trajectory Optimization}
\label{subsec:selection_strategy}

A single rectification update may be insufficient to realize the desired correction.
Repeated updates allow further exploration, but excessive intervention may disrupt the generation trajectory and reduce consistency with the original user instruction.
To address this trade-off, we introduce Greedy Iterative Trajectory Optimization (GITO), which combines iterative exploration with greedy selection to determine the state retained for subsequent denoising.

\noindent\textbf{Iterative Trajectory Exploration.}
Instead of relying on a single update, we explore the local neighborhood of the current latent state through iterative rectification.
For each sampling step within the rectification window $\mathcal{W}=[s_{\mathrm{start}},s_{\mathrm{end}}]$, we perform $K$ ISR updates from $z_t^{(0)}=z_t$ to construct the candidate set $\mathcal{Z}_{\mathrm{cand}}=\{z_t^{(0)},\ldots,z_t^{(K)}\}$.
The first update generates $c_{\mathrm{target}}^{t}$ from $z_t^{(0)}$; subsequent updates at the same timestep reuse this target.
At each iteration, we recompute the look-ahead image and CSA gradient from the current candidate and update the latent as
\begin{equation}
\label{eq:iterative_rectification}
    z_t^{(k+1)}
    =
    z_t^{(k)}
    -
    \eta\,
    \hat{g}_t\!\left(z_t^{(k)}\right),
    \qquad
    k=0,\ldots,K-1.
\end{equation}

\noindent\textbf{Greedy Selection Strategy.}
To avoid over-rectification, we do not assume the final iteration $z_t^{(K)}$ is optimal.
Instead, we employ a Greedy Selection Strategy (GSS) to evaluate all candidates against the original user instruction $c_{\mathrm{user}}$.
The reflective target guides rectification of the current result, while the original instruction provides a common criterion for selection.
This reduces selection’s dependence on potentially inaccurate reflective targets and encourages alignment with the user’s original instruction.
Specifically, for $c_{\mathrm{user}}=(u_1,\ldots,u_J)$, we teacher-force the instruction through the same UMM with each candidate's estimated image as visual context and compute
\begin{equation}
\label{eq:gss_score}
    S_t(z)
    =
    \frac{1}{J}
    \sum_{i=1}^{J}
    \log
    P_\theta
    \!\left(
    u_i
    \mid
    u_{<i},
    \hat{x}_{0|t}(z)
    \right).
\end{equation}
The retained state is selected as
\begin{equation}
\label{eq:gito_selection}
    z_t^*
    =
    \operatorname*{arg\,max}_{z\in\mathcal{Z}_{\mathrm{cand}}}
    S_t(z).
\end{equation}
We resume denoising from the selected state $z_t^*$ and repeat reflection, iterative rectification, and greedy selection at subsequent rectification timesteps.

\section{Experiments}
\label{sec:experiments}

\subsection{Experimental Setup}

\noindent\textbf{Benchmarks and Baselines.}
\label{sec:benchmarks}
To comprehensively assess text-to-image generation capabilities, we utilize GenEval~\cite{ghosh2023genevalobjectfocusedframeworkevaluating} to evaluate compositional generation skills, specifically focusing on object co-occurrence and attribute binding, while employing DPG-Bench~\cite{hu2024ellaequipdiffusionmodels} to test robustness in handling complex and dense prompts.
We apply UniRect-CoT to four representative flow-based UMMs: Janus-Flow~\cite{ma2025janusflowharmonizingautoregressionrectified}, OpenUni-RecA~\cite{xie2025reconstruction}, OmniGen2~\cite{wu2025omnigen2explorationadvancedmultimodal}, and BAGEL~\cite{bagel}.

\noindent\textbf{Implementation Details.}
\label{sec:imp_details}
Unless otherwise specified, we use $50$ flow-matching sampling steps with the default classifier-free guidance settings from each baseline's official implementation.
The rectification window is $\mathcal{W}=[5,10]$ in sampling-step indices, with $K=3$ ISR updates per rectification timestep.
We set the gradient clipping threshold to $\delta=0.001$ and the latent update step size to $\eta=200$.
Evaluations are conducted on NVIDIA H100 GPUs.

\subsection{Quantitative Results}
\label{quantitative}

\definecolor{mygreen}{HTML}{008800}
\definecolor{lightgray}{gray}{0.92}

\begin{table}[t]
    \centering
    \caption{\textbf{Quantitative comparison on GenEval and DPG-Bench.} We compare our method with state-of-the-art unified multimodal models (top) and verify the effectiveness of our method on standard baselines (bottom). Results marked with $\dagger$ are reproduced in this work by generating 4 consecutive samples per prompt, starting with a fixed initial seed. $\uparrow$ denotes that higher is better. We highlight \textbf{improvements} over the baseline and the \underline{best} overall performance.}
    \label{tab:main_table}
    \small 
    
    
    \definecolor{mygreen}{HTML}{008800}
    \newcommand{\cinc}[1]{\rlap{\,\textcolor{mygreen}{\ensuremath{\bm{\uparrow}}}}}
    \resizebox{\linewidth}{!}{
    \begin{tabular}{lcccccccc}
        \toprule
         & DPG $\uparrow$ & \multicolumn{7}{c}{GenEval $\uparrow$} \\
        \cmidrule(lr){2-2} \cmidrule(lr){3-9} 
        Model & Score & Single Obj. & Two Obj. & Counting & Colors & Position & Color Attri. & Overall  \\
        \midrule 
        TokenFlow & 73.4 & 0.97\phantom{0} & 0.66\phantom{0} & 0.40\phantom{0} & 0.84\phantom{0} & 0.17\phantom{0} & 0.26\phantom{0} & 0.55\phantom{0} \\
        OpenUni            & 79.0 & 0.99\phantom{0} & 0.71\phantom{0} & 0.55\phantom{0} & 0.82\phantom{0} & 0.25\phantom{0} & 0.42\phantom{0} & 0.62\phantom{0} \\
        Emu3             & 80.6 & 0.98\phantom{0} & 0.71\phantom{0} & 0.34\phantom{0} & 0.81\phantom{0} & 0.17\phantom{0} & 0.21\phantom{0} & 0.54\phantom{0} \\
        Show-o2           & 85.0  & 0.99\phantom{0} & 0.86\phantom{0} & 0.55\phantom{0} & 0.86\phantom{0} & 0.46\phantom{0} & 0.63\phantom{0} & 0.73\phantom{0} \\
        Janus            & 79.6 & 0.97\phantom{0} & 0.68\phantom{0} & 0.30\phantom{0} & 0.84\phantom{0} & 0.46\phantom{0} & 0.42\phantom{0} & 0.61\phantom{0} \\
        Janus-Pro-7B       & 84.1 & 0.99\phantom{0} & 0.89\phantom{0} & 0.59\phantom{0} & \underline{0.90}\phantom{0} & \underline{0.79}\phantom{0} & 0.66\phantom{0} & {0.80}\phantom{0} \\
        
        \midrule
        \multicolumn{9}{l}{\textit{\textbf{Ours \& Baselines}}} \\
        Janus-Flow $^\dagger$  & 73.0 & 0.950 & 0.482 & 0.346 & 0.771 & 0.352 & 0.277 & 0.530 \\
        
        \rowcolor{lightgray}
        \quad + Ours        & \textbf{75.2}   & \textbf{0.959} & \textbf{0.609} & \textbf{0.419} & \textbf{0.811} & \textbf{0.455} & \textbf{0.350} & \textbf{0.601} \\

        OpenUni-RecA $^\dagger$  & 80.1 & 0.987 & 0.906 & 0.478 & 0.877 & 0.415 & 0.680 & 0.724 \\
        
        \rowcolor{lightgray}
        \quad + Ours   & \textbf{82.5}   & \textbf{0.994} & \textbf{0.934} & \textbf{0.538} & \textbf{0.891} & \textbf{0.423} & \underline{\textbf{0.698}} & \textbf{0.746} \\
        
        OmniGen2 $^\dagger$  & 83.0  & 0.993 & 0.929 & 0.728 & 0.869 & 0.535 & 0.665 & 0.786 \\
        
        \rowcolor{lightgray}
        \quad + Ours        & \textbf{84.2}   & \textbf{0.994} & \textbf{0.957} & 0.728 & \textbf{0.875} & \textbf{0.553} & {\textbf{0.670}} & \textbf{0.796} \\

        BAGEL $^\dagger$    & {85.0}   & 0.991 & 0.957 & 0.743 & 0.867 & 0.490 & 0.610 & 0.776 \\
        
        \rowcolor{lightgray}
        \quad + Ours         & \underline{\textbf{85.9}}  & \underline{\textbf{0.997}} & \underline{\textbf{0.970}} & \underline{\textbf{0.819}} & \textbf{0.872} & 0.485 & \textbf{0.680} & \underline{\textbf{0.804}} \\
        
        
        \bottomrule
    \end{tabular}
        }
    \vspace{-3mm}
\end{table}

\noindent\textbf{Comparison across Unified Multimodal Models.}
As shown in \Cref{tab:main_table}, UniRect-CoT improves the overall scores of all four baselines on both benchmarks.
Janus-Flow exhibits the largest GenEval gain, from $0.530$ to $0.601$, while its DPG-Bench score increases from $73.0$ to $75.2$.
On the stronger BAGEL baseline, UniRect-CoT improves GenEval Overall from $0.776$ to $0.804$, with notable gains in counting ($0.743 \rightarrow 0.819$) and attribute binding ($0.610 \rightarrow 0.680$).
These results support the applicability of trajectory rectification across UMMs with different architectures and baseline strengths.

\noindent\textbf{Comparison with Test-Time Reasoning and Refinement.}
As shown in \Cref{tab:comparison}, UniRect-CoT achieves the highest GenEval Overall score among the evaluated training-free methods on BAGEL.
It improves the baseline from $77.6$ to $80.4$, outperforming Qwen Rewrite~\cite{bai2025qwen3vltechnicalreport} ($78.4$), UiG~\cite{lyu2025understandingingenerationreinforcinggenerativecapability} ($78.1$), and TiR~\cite{khan2025testtimepromptrefinementtexttoimage} ($79.8$).
As shown in \Cref{fig:ablation_efficiency}(c), UniRect-CoT
achieves this score at $1.74\times$ the latency of standard
$50$-step BAGEL, with lower latency than both UiG and TiR.
Qwen Rewrite is faster but achieves a lower score of $78.4$,
highlighting the performance--latency trade-off.
We also evaluate BAGEL's Think mode, whose pre-generation reasoning capability is trained from reasoning aligned data~\cite{bagel}.
Think improves the baseline score to $79.7$, and applying UniRect-CoT raises it to $81.3$ without additional training.
This gain highlights the complementarity between learned pre-generation reasoning and our training-free trajectory rectification.

\providecommand{\cmark}{\ding{51}}
\providecommand{\xmark}{\ding{55}}

\begin{table*}[t]
\centering
\caption{
\textbf{Comparison of test-time strategies on BAGEL.}
``Pre'': pre-generation reasoning or prompt enhancement;
``Multi'': multiple full generation/refinement rounds;
``Traj.'': intervention during denoising;
``State'': feedback conditioned on the current visual result.
$\dagger$ marks results reproduced under our unified inference and evaluation protocol.
Boldface highlights results with UniRect-CoT.
}
\label{tab:comparison}
\small
\setlength{\tabcolsep}{3.2pt}
\renewcommand{\arraystretch}{1.13}
\resizebox{\textwidth}{!}{%
\begin{tabular}{@{}lccccccccccc@{}}
\toprule
\multicolumn{5}{c}{\textbf{Configuration}} &
\multicolumn{7}{c}{\textbf{GenEval} $\uparrow$} \\
\cmidrule(lr){1-5}
\cmidrule(lr){6-12}
\textbf{Method}
& \textbf{Pre}
& \textbf{Multi}
& \textbf{Traj.}
& \textbf{State}
& \textbf{Single Obj.}
& \textbf{Two Obj.}
& \textbf{Counting}
& \textbf{Colors}
& \textbf{Position}
& \textbf{Color Attri.}
& \textbf{Overall} \\
\midrule
\multicolumn{12}{@{}l@{}}{
    \textit{Training-free test-time scaling on BAGEL}
} \\[-1pt]
BAGEL
& \xmark & \xmark & \xmark & \xmark
& 99.1 & 95.7 & 74.3 & 86.7 & 49.0 & 61.0 & 77.6 \\
Qwen Rewrite $\dagger$
& \cmark & \xmark & \xmark & \xmark
& 100.0 & 95.7 & 73.4 & 89.9 & 47.3 & 64.0 & 78.4 \\
UiG $\dagger$
& \cmark & \cmark & \xmark & \cmark
& 99.1 & 95.7 & 75.3 & 87.8 & 48.8 & 61.5 & 78.1 \\
TiR $\dagger$
& \cmark & \cmark & \xmark & \cmark
& 98.4 & 95.7 & 80.0 & 83.5 & 54.0 & 67.2 & 79.8 \\
\textbf{UniRect-CoT}
& \xmark & \xmark & \cmark & \cmark
& \textbf{99.7}
& \textbf{97.0}
& \textbf{81.9}
& \textbf{87.2}
& \textbf{48.5}
& \textbf{68.0}
& \textbf{80.4} \\
\midrule
\multicolumn{12}{@{}l@{}}{
    \textit{Test-time scaling on reasoning-tuned BAGEL}
} \\[-1pt]
BAGEL Think $\dagger$
& \cmark & \xmark & \xmark & \xmark
& 99.1 & 93.6 & 77.8 & 87.2 & 54.0 & 66.7 & 79.7 \\
\textbf{Think + UniRect-CoT}
& \cmark & \xmark & \cmark & \cmark
& \textbf{99.4}
& \textbf{93.9}
& \textbf{79.4}
& \textbf{89.4}
& \textbf{54.5}
& \textbf{70.8}
& \textbf{81.3} \\
\bottomrule
\end{tabular}%
}
\vspace{-3mm}
\end{table*}

\begin{figure*}[h!]
    \centering
    \includegraphics[width=0.95\textwidth]{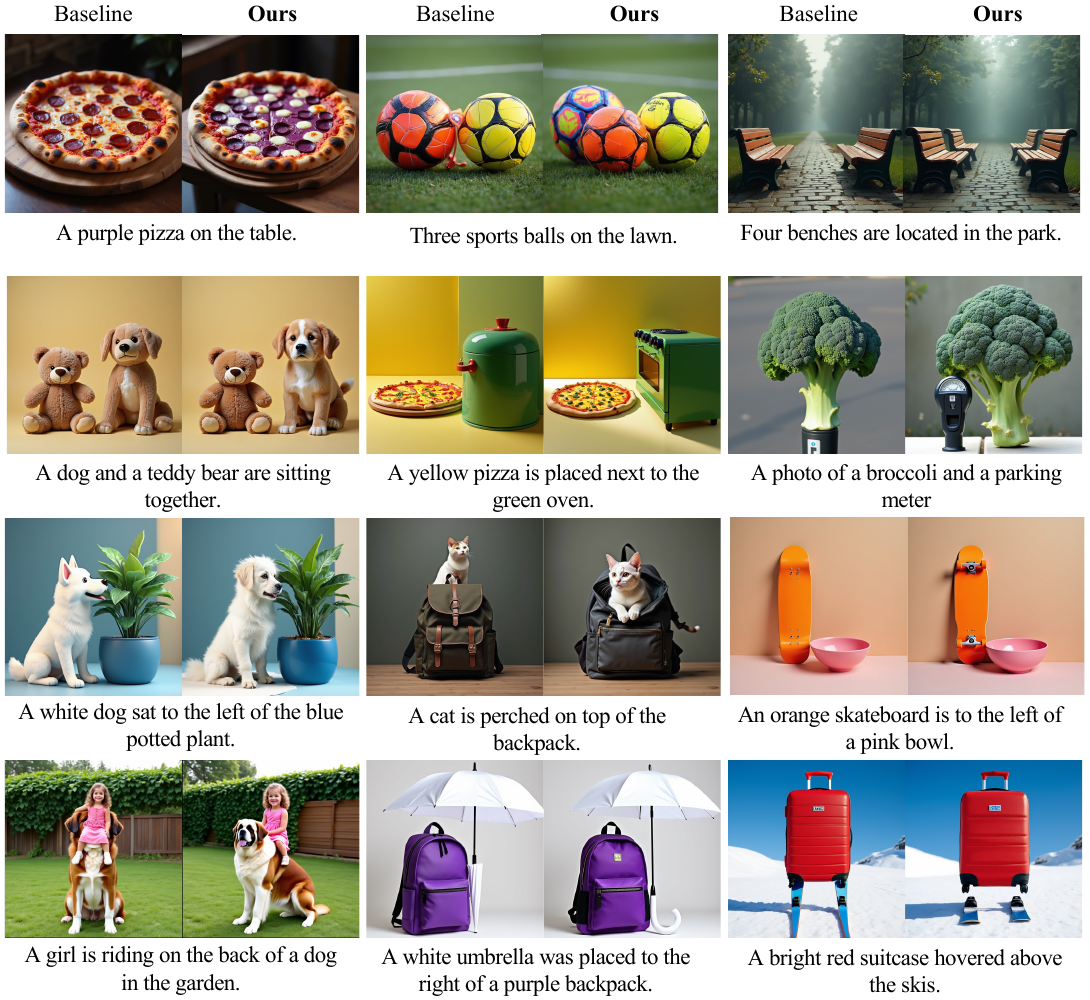}
    \caption{
    \textbf{Qualitative comparison.}
    Each pair shows BAGEL (left) and UniRect-CoT (right).
    }
    \label{fig:cmp}
    \vspace{-5mm}
\end{figure*}

\subsection{Qualitative Comparison}

\Cref{fig:cmp} presents representative comparisons with BAGEL on challenging compositional prompts.
In these examples, UniRect-CoT more accurately follows the requested object counts, colors, attribute assignments, and spatial relations, illustrating its ability to correct compositional errors.

\subsection{Ablation Studies and Efficiency Analysis}
\label{sec:ablation}

We conduct ablation studies on BAGEL using GenEval to evaluate rectification strategy, efficiency, and semantic target design. 

\begin{figure*}[t]
\centering
\begin{subfigure}[t]{0.32\textwidth}
    \centering
    \includegraphics[width=\linewidth]{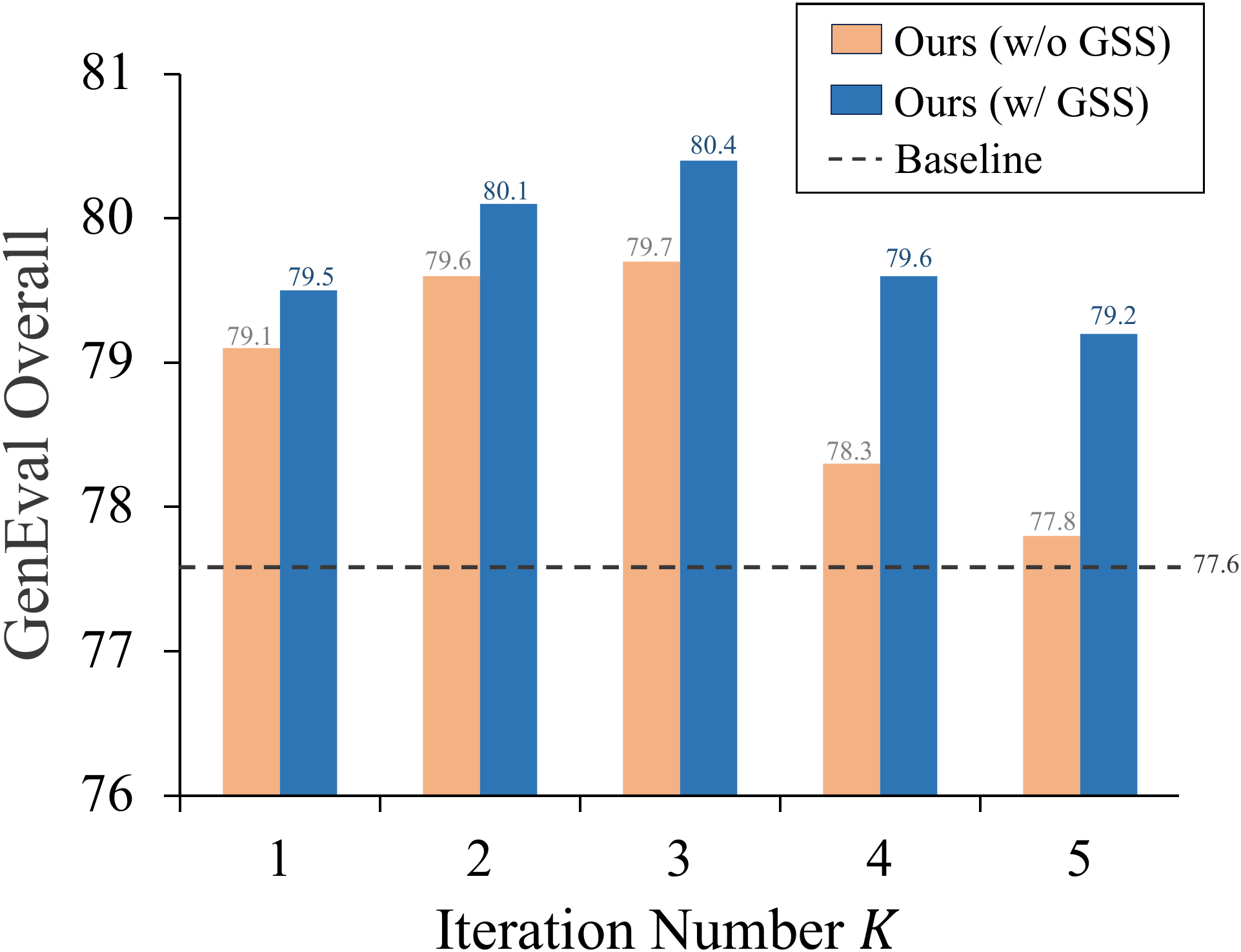}
    \caption{Iterative Rectification}
    \label{fig:ablation_iter}
\end{subfigure}
\hfill
\begin{subfigure}[t]{0.32\textwidth}
    \centering
    \includegraphics[width=\linewidth]{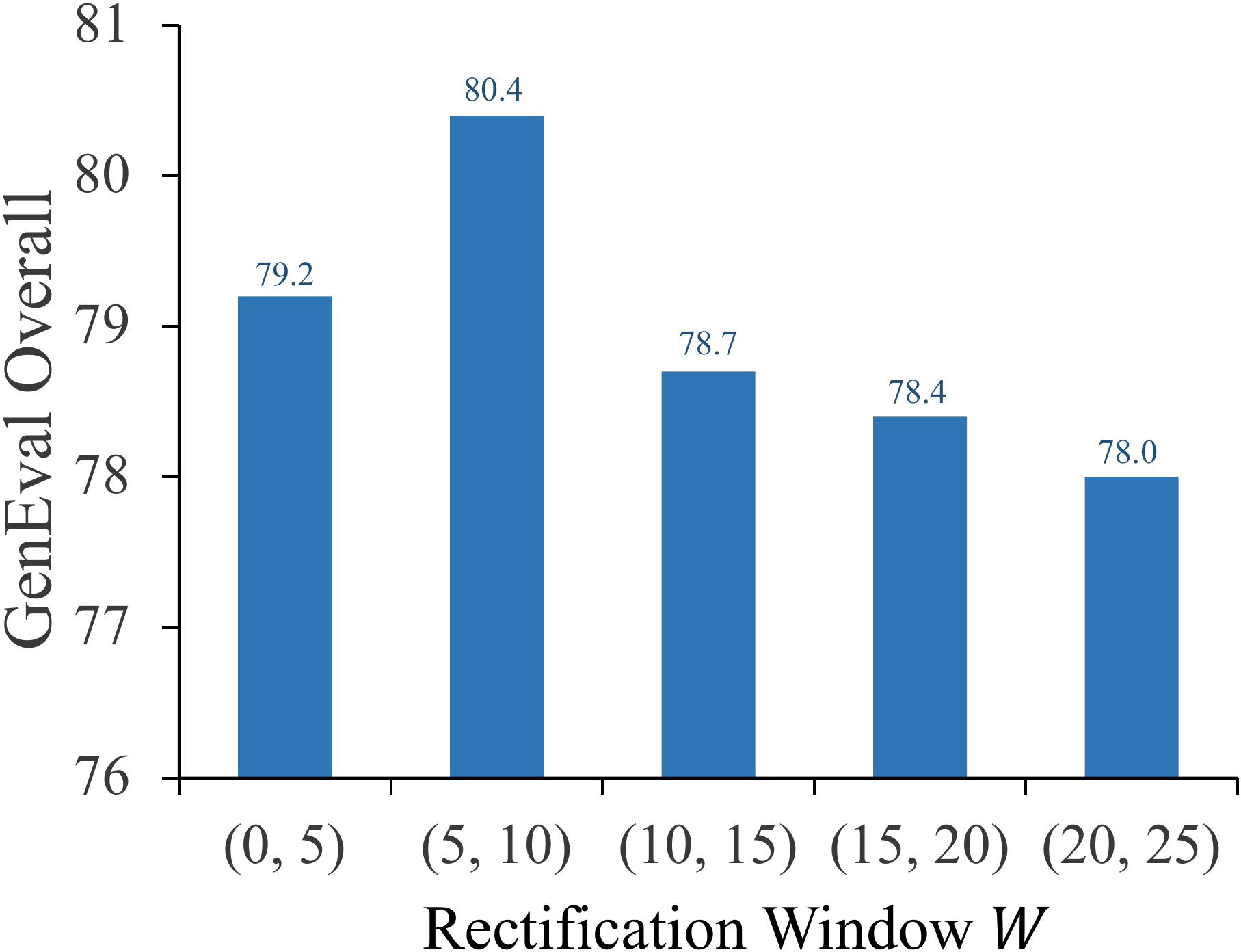}
    \caption{Rectification Window}
    \label{fig:ablation_window}
\end{subfigure}
\hfill
\begin{subfigure}[t]{0.32\textwidth}
    \centering
    \includegraphics[width=\linewidth]{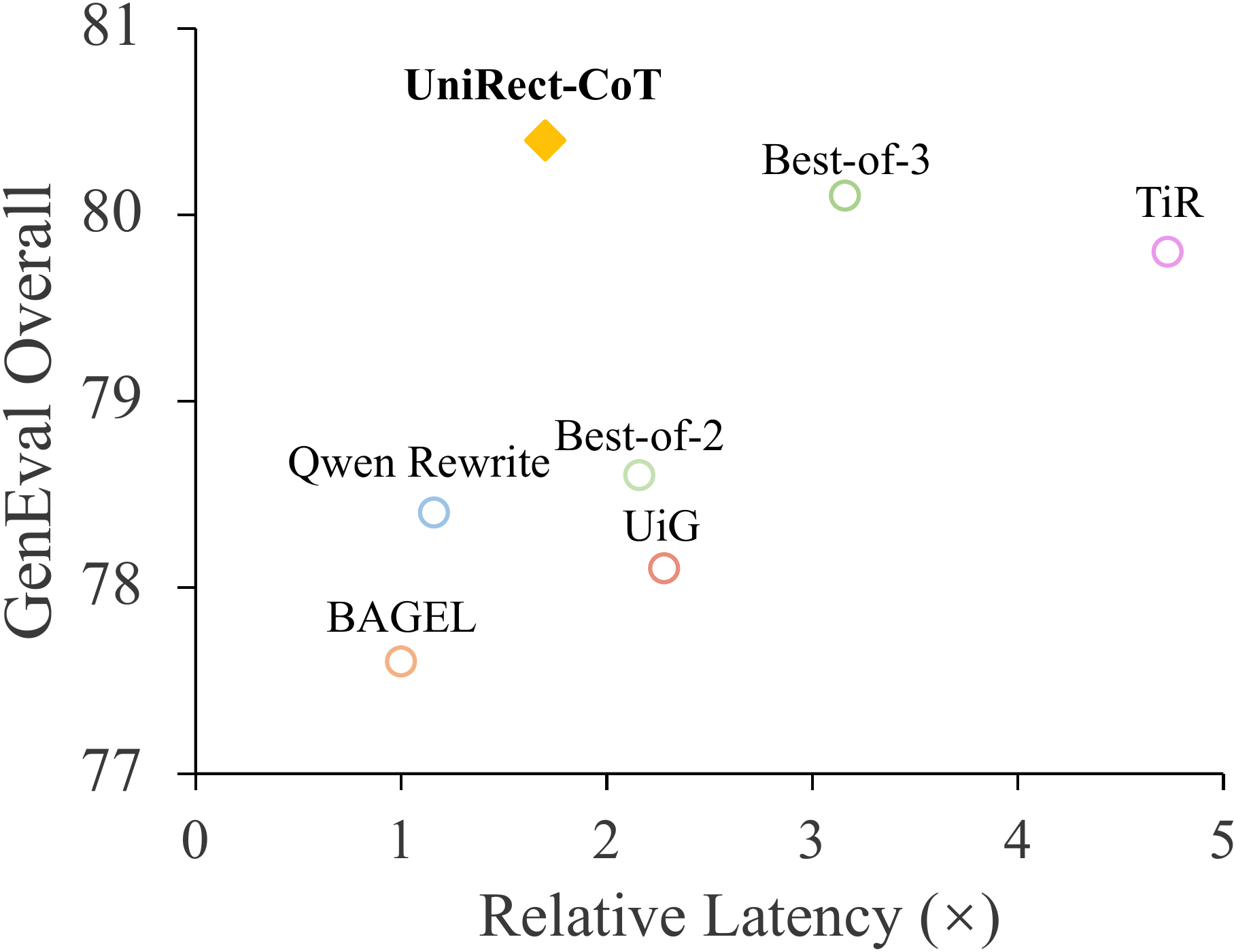}
    \caption{Performance vs.\ Latency}
    \label{fig:efficiency}
\end{subfigure}

\caption{
\textbf{Ablation of rectification strategy and efficiency Analysis.}
(a) Effect of the number of rectification updates $K$ and the Greedy Selection Strategy (GSS).
(b) Effect of the rectification window $\mathcal{W}$ at $K=3$.
(c) GenEval Overall versus relative latency for UniRect-CoT and competing test-time strategies.
Latency is normalized to standard BAGEL inference ($1.0\times$, 24.7 s/image).
}
\label{fig:ablation_efficiency}
\end{figure*}

\noindent\textbf{Rectification Strategy and Efficiency Analysis. }
\Cref{fig:ablation_efficiency}(a) shows that performance peaks at $K=3$, reaching $80.4$ with GSS and $79.7$ without GSS, and declines with further updates.
GSS consistently improves performance and mitigates over-rectification: at $K=5$, it achieves $79.2$, compared with $77.8$ without selection, although both remain below their scores at $K=3$.
We therefore use $K=3$ to balance quality and cost.
The best window is $\mathcal{W}=[5,10]$ (\Cref{fig:ablation_efficiency}(b)); earlier and later windows yield lower scores, suggesting a balance between reflection reliability and the flexibility of the intermediate state.
\Cref{fig:ablation_efficiency}(c) further compares UniRect-CoT with alternative test-time strategies.
UniRect-CoT achieves higher GenEval Overall scores with lower latency than the evaluated Best-of-$N$ configurations.
Qwen Rewrite incurs less latency but achieves a lower score,
highlighting a different trade-off.

\noindent\textbf{Semantic Target Design.}
\Cref{tab:target_ablation} examines whether reflection provides better guidance than directly using the original instruction as the target for gradient-based latent updates.
Our method improves Overall from $78.7$ to $79.7$ without GSS and from $79.1$ to $80.4$ with GSS compared with user-only guidance.
In contrast, text-only expansion underperforms the original instruction in both settings, suggesting that the gains come from feedback on the current image rather than prompt expansion alone.
Static reflection already improves upon the original instruction, and dynamic reflection provides further gains, supporting the value of updating the corrective target as the visual state evolves.
For candidate selection, however, reflection-based GSS achieves $78.8$, below both instruction-based GSS ($80.4$) and no selection ($79.7$).
These results support distinct roles for guidance and selection: reflection adapts the proposed correction to the current image, while the original instruction provides a consistent criterion for evaluating candidates.
Building on the general guidance mechanism of UG~\cite{bansal2023universal}, our method introduces adaptive targets generated through the UMM's own understanding and a separate selection criterion based on the original instruction.
The ablation results demonstrate the benefit of these choices over directly optimizing a fixed user instruction.

\begin{table*}[t]
\centering
\caption{
\textbf{Ablation of semantic target design.}
``User'', ``Exp.'', and ``Ref.'' denote the original instruction $c_{\mathrm{user}}$, its text-only expansion, and the state-conditioned reflective target $c_{\mathrm{target}}^{t}$, respectively.
Static targets remain fixed across the rectification window; dynamic targets are regenerated at each rectification timestep and reused for its $K$ updates.
``--'' indicates that the corresponding component is disabled.
}
\label{tab:target_ablation}
\small
\setlength{\tabcolsep}{3.2pt}
\renewcommand{\arraystretch}{1.13}
\resizebox{\textwidth}{!}{%
\begin{tabular}{@{}clllccccccc@{}}
\toprule
\multicolumn{4}{c}{\textbf{Configuration}} &
\multicolumn{7}{c}{\textbf{GenEval} $\uparrow$} \\
\cmidrule(lr){1-4}
\cmidrule(lr){5-11}
\textbf{Setting} &
\textbf{Update} &
\textbf{ITE Target} &
\textbf{GSS Target} &
\textbf{Single Obj.} &
\textbf{Two Obj.} &
\textbf{Counting} &
\textbf{Colors} &
\textbf{Position} &
\textbf{Color Attri.} &
\textbf{Overall} \\
\midrule
\textsc{Baseline}
& -- & -- & --
& 99.1 & 95.7 & 74.3 & 86.7 & 49.0 & 61.0 & 77.6 \\
\midrule
\multirow{4}{*}{\textsc{ITE only}}
& Static & User & --
& 99.7 & 93.9 & 76.3 & 86.7 & 48.5 & 67.0 & 78.7 \\
& Static & Exp. & --
& 99.7 & 94.2 & 76.9 & 87.2 & 48.8 & 60.5 & 77.9 \\
& Static & Ref. & --
& 100.0 & 94.7 & 77.8 & 86.2 & 46.5 & 69.3 & 79.1 \\
& Dynamic & Ref. & --
& 100.0 & 95.5 & 80.3 & 87.8 & 46.0 & 69.0 & 79.7 \\
\midrule
\multirow{5}{*}{\textsc{ITE + GSS}}
& Static & User & User
& 99.4 & 95.2 & 76.5 & 88.3 & 51.0 & 64.3 & 79.1 \\
& Static & Exp. & User
& 99.4 & 94.4 & 77.8 & 87.5 & 48.8 & 60.5 & 78.1 \\
& Static & Ref. & User
& 99.7 & 96.2 & 77.2 & 86.2 & 47.8 & 69.3 & 79.4 \\
& Dynamic & Ref. & Ref.
& 99.1 & 95.7 & 79.4 & 85.4 & 47.0 & 66.5 & 78.8 \\
& Dynamic & Ref. & User
& 99.7 & 97.0 & 81.9 & 87.2 & 48.5 & 68.0 & 80.4 \\
\bottomrule
\end{tabular}%
}
\end{table*}
\section{Conclusion}
\label{sec:conclusion}
In this paper, we proposed UniRect-CoT, a training-free unified rectification chain-
of-thought framework that uses the inherent understanding of UMMs to improve image generation.
Inspired by ``Thinking-While-Drawing,'' our framework interleaves reflection and rectification within a single denoising trajectory.
Intrinsic Semantic Rectification converts state-conditioned corrective targets into gradients for latent updates, while Greedy Iterative Trajectory Optimization selects among the resulting candidates using the original user instruction.
Extensive experiments on four flow-based UMMs show consistent improvements on GenEval and DPG-Bench, demonstrating the effectiveness and generality of UniRect-CoT.

\newpage

\bibliography{main}
\bibliographystyle{iclr2027_conference}

\end{document}